%% file: main.tex
\documentclass{article}

\usepackage[final]{neurips_2024}

\usepackage[utf8]{inputenc} 
\usepackage[T1]{fontenc}    
\usepackage{hyperref}       
\usepackage{url}            
\usepackage{booktabs}       
\usepackage{amsfonts}       
\usepackage{nicefrac}       
\usepackage{microtype}      
\usepackage{multirow}
\usepackage{pifont}
\usepackage{threeparttable}
\usepackage{amsmath}
\usepackage{graphicx}
\usepackage{subfigure}
\usepackage{wrapfig}
\usepackage[table,xcdraw]{xcolor}
\usepackage{floatrow}
\usepackage{caption}
\newfloatcommand{figurebox}{figure}[\nocapbeside][\dimexpr(\textwidth-\columnsep)/2\relax]
\newfloatcommand{tablebox}{table}[\nocapbeside][\dimexpr(\textwidth-\columnsep)/2\relax]
\title{DN-4DGS: Denoised Deformable Network with Temporal-Spatial Aggregation for Dynamic Scene Rendering}
\definecolor{yzybest}{rgb}{0.96, 0.57, 0.58}
\definecolor{yzysecond}{rgb}{0.98, 0.78, 0.57}
\definecolor{yzythird}{rgb}{1.0, 1.0, 0.56}

\author{%
Jiahao Lu$^{1}$~\quad Jiacheng Deng$^{1}$ ~\quad Ruijie Zhu$^{1}$ ~\quad Yanzhe Liang$^{1}$ ~\quad 
\\ \textbf{Wenfei Yang}$^{1}$ ~\quad \textbf{Tianzhu Zhang}$^{1,2}$\thanks{Corresponding author}~\quad \textbf{Xu Zhou}$^{3}$ \\\\\
$^1$University of Science and Technology of China, $^2$Deep Space Exploration Lab, \\$^3$Sangfor Technologies Inc.\\
\texttt{\{lujiahao, dengjc, ruijiezhu, yzliang\}@mail.ustc.edu.cn,}\\ 
\texttt{\{yangwf, tzzhang\}@ustc.edu.cn, zhouxu@sangfor.com.cn} \\
}

\begin{document}

\maketitle
\begin{figure}[!ht]
  \begin{center}
      \includegraphics[width=0.95\textwidth]{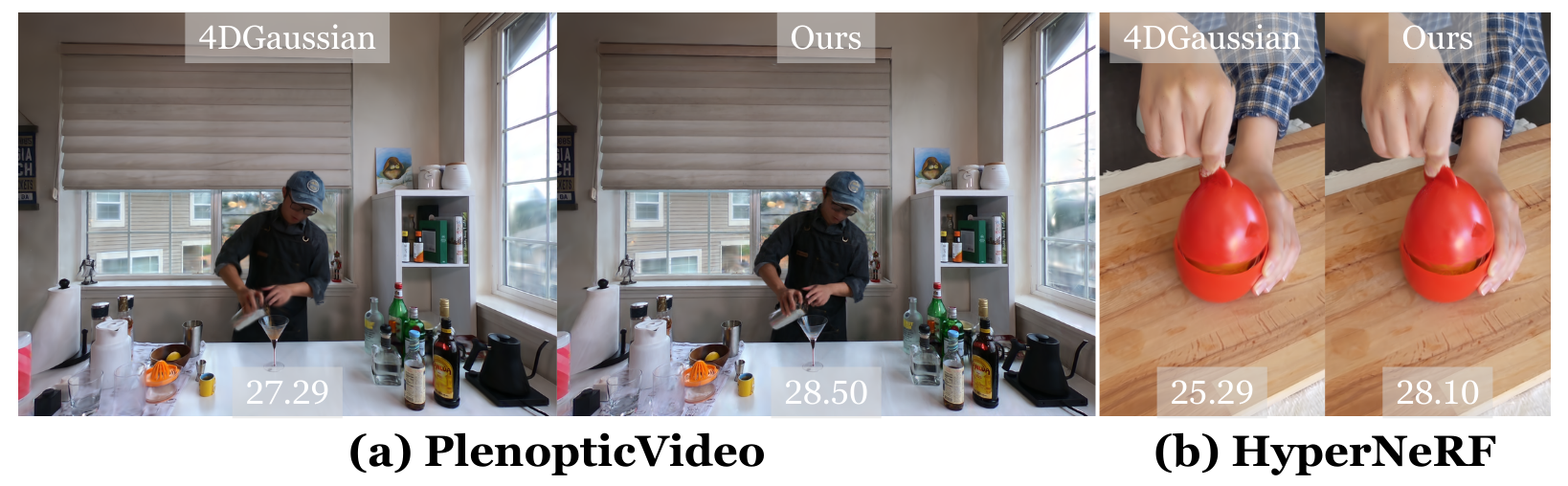}
      \caption{(a) The visualization results on PlenopticVideo~\cite{li2022neural} dataset.  (b) The visualization results on HyperNeRF~\cite{park2021hypernerf} dataset. The numbers below the images represent PSNR.}
      \label{intro1}
  \end{center}
  \vspace{-1.3em}
\end{figure}
\begin{abstract}
\input{abst}
\end{abstract}

\vspace{-0.7em}
\section{Introduction}
\vspace{-5pt}
\input{intro}

\vspace{-5pt}
\vspace{-0.4em}
\section{Related Work}
\vspace{-5pt}
\input{rw}

\section{Preliminary: 3D Gaussian Splatting}
\vspace{-5pt}
\label{preliminary}
\input{preliminary}

\vspace{-5pt}
\vspace{-0.2em}
\section{Method}
\label{Method}
\vspace{-5pt}
\input{method}

\vspace{-5pt}
\section{Experiment}
\label{Experiment}
\input{expr}

\vspace{-5pt}
\section{Conclusion}
\vspace{-5pt}
\input{clu}

{
\small
\bibliographystyle{unsrt}
\bibliography{egbib}
}
\appendix

\section{Appendix / supplemental material}
\input{supplementary}

\end{document}

%% file: abst.tex
Dynamic scenes rendering is an intriguing yet challenging problem. 
Although current methods based on NeRF have achieved satisfactory performance, they still can not reach real-time levels. 
Recently, 3D Gaussian Splatting (3DGS) has garnered researchers' attention due to their outstanding rendering quality and real-time speed. 
Therefore, a new paradigm has been proposed: defining a canonical 3D gaussians and deforming it to individual frames in deformable fields.
However, since the coordinates of canonical 3D gaussians are filled with noise, which can transfer noise into the deformable fields, and there is currently no method that adequately considers the aggregation of 4D information.
Therefore, we propose Denoised Deformable Network with Temporal-Spatial Aggregation for Dynamic Scene Rendering (DN-4DGS).
Specifically, a Noise Suppression Strategy is introduced to change the distribution of the coordinates of the canonical 3D gaussians and suppress noise. 
Additionally, a Decoupled Temporal-Spatial Aggregation Module is designed to aggregate information from adjacent points and frames.
Extensive experiments on various real-world datasets demonstrate that our method achieves state-of-the-art rendering quality under a real-time level. Code is available at \href{https://github.com/peoplelu/DN-4DGS}{https://github.com/peoplelu/DN-4DGS}.

%% file: intro.tex
\label{sec:intro}
Dynamic scene reconstruction from single or multi-view videos is a crucial task in computer vision, with applications such as VR/AR~\cite{kim2019immersive,pages2018affordable}, 3D perception~\cite{ lu2024bsnet,li2024mamba24,deng2024diff3detr}, movie production~\cite{hisatomi2009method}, etc. Neural Radiance Fields (NeRF)~\cite{mildenhall2021nerf} offer a promising approach by representing scenes with implicit functions derived from multi-view inputs. By incorporating time as an additional input~\cite{pumarola2021dnerf,li2022neural}, NeRF enables dynamic scene rendering. However, the original NeRF model suffers from significant training and rendering costs, attributed to the high number of points sampled per camera ray and volume rendering.

\begin{figure}[!t]
    \vspace{-2.0em}
    \begin{center}
        \includegraphics[width=0.95\textwidth]{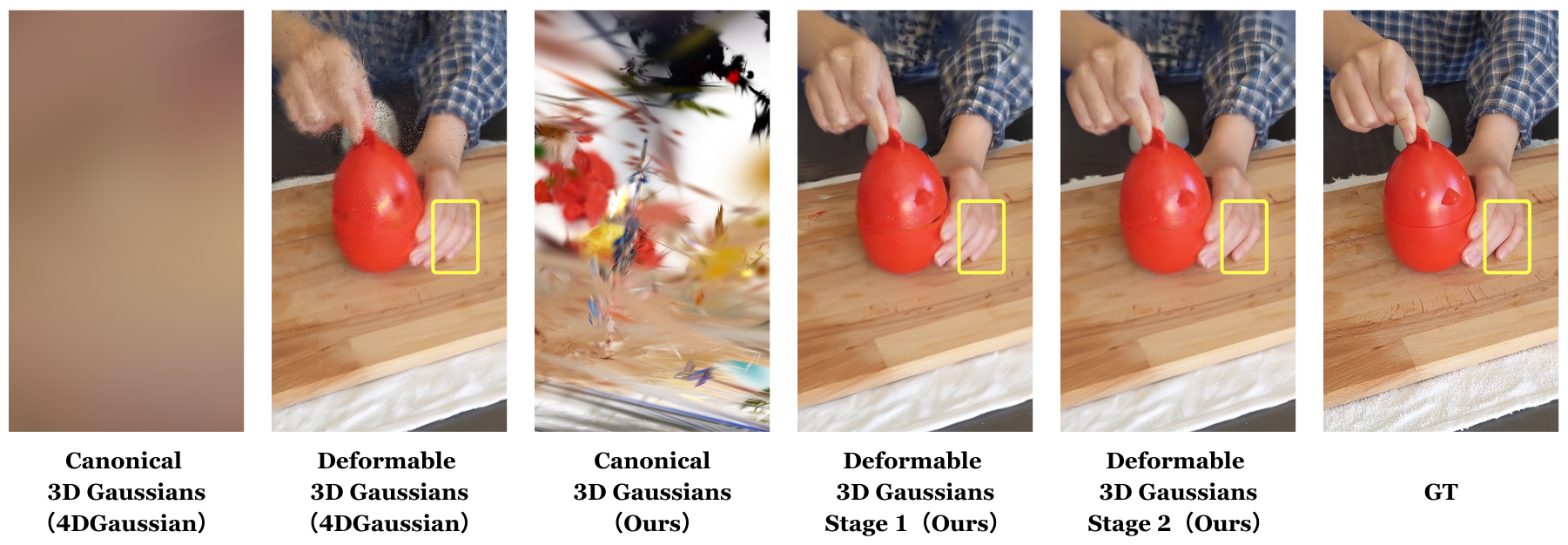}
        \caption{\textbf{Comparison of our render visualization with 4DGaussian~\cite{wu20234d}. }The results are rendered on HyperNeRF~\cite{park2021hypernerf} dataset and use the point cloud provided by HyperNeRF for Gaussian initialization ($Sparse$ $Init$). Image 1: canonical 3D gaussians generated by 4DGaussian. Image 2: deformable 3D gaussians generated by 4DGaussian. Image 3: canonical 3D gaussians generated by our method. Image 4: deformable 3D gaussians after the first stage. Image 5: deformable 3D gaussians after the second stage. Image 6: ground truth. The yellow box emphasizes that through a two-stage deformation process, our method can produce higher-quality rendering results.}
        \label{motivation}
        \vspace{-1.3em}
    \end{center}
\end{figure}
Recently, the emerging 3D Gaussian Splatting (3DGS)~\cite{3dgs} has significantly increased rendering speed to a real-time level compared to NeRFs by employing a differentiable rasterizer for 3D Gaussian primitives. 3DGS directly optimizes the parameters of 3D gaussians (position, opacity, anisotropic covariance, and spherical harmonics (SH) coefficients) and renders them through projection and $\alpha $-blending. Given the explicit expression nature of 3DGS, recent studies~\cite{wu20234d,yang2023deformable,huang2023sc} represent dynamic scenes by defining a canonical 3D gaussians and deforming it to individual frames in deformable fields. Specifically, during the execution of the deformation field, the coordinates $xyz$ along with time $t$ are used as input, and the output corresponds to the changes in Gaussian properties. It is noteworthy that the existing methods, when designing deformable networks, either directly map the 4D coordinates of each input point to the latent space using MLPs~\cite{yang2023deformable,huang2023sc}, or use HexPlane~\cite{hexplane} to interpolate a series of learnable embeddings to obtain the latent of each point~\cite{wu20234d}. 

Both of these approaches have drawbacks that can not be ignored.
1) Canonical 3D gaussians are synthesized from multi-frame images of dynamic scenes. 
Due to the presence of dynamic regions and the specific design of A (canonical 3D gaussians) + B (deformable network), canonical 3D gaussians exhibit significant noise, as illustrated in Figure~\ref{motivation}.
This noise is inevitably transferred to the deformable field after the input $xyz$ is passed through the deformable network. To elaborate on the ``Noise'': In the canonical + deformable design, we input the canonical Gaussian coordinates $xyz$ and time $t$ into the deformable network. The deformable network essentially performs basic operations (addition, subtraction, multiplication, division) on the coordinates $xyz$ and time $t$. Since the point-to-point relationships within the canonical Gaussians are chaotic and erroneous, as shown in Figure~\ref{motivation}, it is predictable that feeding these erroneous coordinates into the deformable network will transfer this error into the deformation field, introducing inaccuracies in the final deformations $\Delta x, \Delta y, \Delta z$.
2) There is a lack of feature aggregation for spatial-temporal information, yet due to the presence of noise in canonical 3D gaussians' $xyz$, direct feature aggregation for spatial information would further amplify noise, affecting the learning of the deformable field. Therefore, spatial aggregation after denoising is very crucial.

To address the aforementioned issues, we propose Denoised Deformable Network with Temporal-Spatial Aggregation for Dynamic Scene Rendering (DN-4DGS), 
primarily consisting of two components: the Noise Suppression Strategy (NSS) and the Decoupled Temporal-Spatial Aggregation Module (DTS). 
To address the initial issue, the design of NSS incorporates two deformation operations. 
The first deformation operation is a standard deformation. It takes the coordinates $xyz$ of the canonical 3D gaussians and time $t$ as input and outputs corresponding coordinate deformations $\Delta x, \Delta y, \Delta z$.
The second deformation builds upon the first by adding $\Delta x, \Delta y, \Delta z$ to the original $xyz$, creating a modified set of coordinates that is then input into a new feature extraction network. The entire process is illustrated by image 3, 4 and 5 in Figure~\ref{motivation}.
This strategy achieves a successful alteration of the distribution of coordinates $xyz$ through the initial coordinate deformation, resulting in noise reduction and the generation of a more accurate deformation field.
It's worth noting that during the early stage of training, we only perform the first deformation operation. 
Only after achieving acceptable results with the first deformation operation do we proceed to the second deformation operation to further enhance accuracy.
To address the second problem, we design the DTS. 
The reason we decouple spatial-temporal aggregation is due to the presence of noise in the coordinates of the canonical 3D gaussians. 
If we directly perform spatial aggregation on the coordinates of the canonical 3D gaussians, the noise information is inevitably amplified after a series of aggregation operations such as k-nearest neighbors (KNN)~\cite{qi2017pointnet++}, significantly affecting the results of the deformation field. 
Therefore, based on the first design NSS, we conduct spatial aggregation during the second deformation operation. 
Considering that temporal information is unrelated to the canonical 3D gaussians, temporal aggregation can be directly incorporated into the first deformation operation to enhance feature extraction capabilities. 
In order to reduce computational overhead and considering that temporal information has already been effectively extracted in the first deformation operation, we do not perform temporal aggregation in the second deformation operation.
In conclusion, our main contributions are outlined as follows:

(i) We introduce a novel representation called Denoised Deformable Network with Temporal-Spatial Aggregation for high-fidelity and efficient dynamic scene rendering.

(ii) We promose the Noise Suppression Strategy, which can change the distribution of the coordinates of the canonical 3D gaussians, suppress noise and generate a more precise deformation field.

(iii) We promose the Decoupled Temporal-Spatial Aggregation Module to aggregate information from adjacent points and frames.

(iv) Extensive experiments on various real-world datasets demonstrate that our method achieves state-of-the-art rendering quality under a real-time level.

%% file: rw.tex
{\bf Dynamic NeRF.}
Novel view synthesis has been a hot topic in academia for several years. NeRF~\cite{mildenhall2021nerf} models static scenes implicitly using MLPs, and numerous studies~\cite{guo2023forward,li2021neural,park2021nerfies,park2021hypernerf,pumarola2021dnerf,tretschk2021non,xian2021space} have extended its application to dynamic scenes through a canonical 3D grid structure and a deformation field. HyperNeRF~\cite{park2021hypernerf} models object topology deformation using higher-dimensional inputs, while DyNeRF~\cite{li2022neural} employs time-conditioned NeRF to represent a 4D scene. However, these approaches, based on vanilla NeRF, suffer from high computational costs due to ray point sampling and volume rendering.
To address this issue, several acceleration methods~\cite{liu2022devrf,li2023dynibar,lin2022efficient,lin2023im4d,lombardi2021mixture,peng2023representing,hexplane,fang2022fast,fridovich2023k,tensor4d,wang2023mixedvoxels,wang2023neural} have been proposed for rendering dynamic scenes. DeVRF~\cite{liu2022devrf} introduces a grid representation, while IBR-based methods~\cite{lin2022efficient,lin2023im4d} utilize multi-camera information for improved quality and efficiency. TensorRF~\cite{chen2022tensorf} adopts multiple planes as explicit representations for direct dynamic scene modeling. Recent approaches such as K-Planes~\cite{kplanes}, Tensor4D~\cite{tensor4d}, and HexPlane~\cite{hexplane} have also been proposed. NeRFPlayer~\cite{song2023nerfplayer} introduces a unified streaming representation for both grid-based and plane-based methods, utilizing separate models to differentiate static and dynamic scene components, albeit at the cost of slow rendering times. HyperReel~\cite{attal2023hyperreel} suggests a flexible sampling network coupled with two planes for dynamic scene representation. Despite the improvements in training and rendering speed achieved by these methods, they still fall short of meeting real-time requirements.

{\bf Dynamic Gaussian Splatting.}
Recently, 3D Gaussian Splatting (3DGS)~\cite{3dgs} has garnered increasing attention from researchers due to its superior rendering quality and real-time rendering speed. 
The method employs a soft point representation with attributes including position, rotation, density, and radiance, and utilizes differentiable point-based rendering for scene optimization. 
Soon after, several concurrent works~\cite{wu20234d,yang2023deformable,huang2023sc,zhu2024motiongs}  have adapted 3D Gaussians for dynamic scenes. 
These methods represent dynamic scenes by establishing a canonical 3DGS and deforming it to individual frames using deformable fields. 
Yang et al.~\cite{yang2023deformable}  predict per-Gaussian offsets using an additional MLP on canonical 3D gaussians, while Wu et al.~\cite{wu20234d}  substitute the MLP with multi-resolution HexPlanes~\cite{hexplane} and a lightweight MLP. 
Our work introduces the Noise Suppression Strategy to change the distribution of the coordinates of the canonical 3D gaussians and generate a more precise deformation field. 
Additionally, for better aggregation of temporal-spatial information, we propose the Decoupled Temporal-Spatial Aggregation Module to consolidate information from adjacent points and frames.

%% file: preliminary.tex
Given images at multiple known viewpoints and timesteps, 3D Gaussian Splatting (3DGS)~\cite{3dgs} optimizes a set of attributes (position, opacity, anisotropic covariance and spherical harmonics) via differentiable rasterization. 3DGS can realize high-fidelity rendering of static 3D scenes in real-time.

Suppose a 3D Gaussian $G(i)$ has the following attributes: position $\mu_i$, opacity $\sigma_i $, covariance matrix $\Sigma_i$ and spherical harmonics $h_i$. The covariance matrix $\Sigma_i$ is decomposed as $\Sigma_i = \mathcal{R} \mathcal{S} \mathcal{S}^T \mathcal{R}^T $
for optimization, with $\mathcal{R}$  as a rotation matrix represented by
a quaternion $q\in \mathbf{S} \mathbf{O} (3)$, and $\mathcal{S}$ as a scaling matrix represented by a 3D vector $s$. Each Gaussian has an opacity value
$\sigma_i $ to adjust its influence in rendering and is associated with
sphere harmonic (SH) coefficients $h_i$ for view-dependent appearance. The final opacity of a 3D gaussian at any spatial point x can be represented as:
\begin{equation}
  \label{opacity}
  \alpha_i= \sigma_i e^{-\frac{1}{2}(x-\mu_i)^T \Sigma_i^{-1}(x-\mu_i)}.
\end{equation}

To render a 2D image, 3D gaussians are projected to 2D space and aggregating them using fast
$\alpha $-blending. The 2D covariance matrix and center are $\Sigma_i^{2D} = JW\Sigma W^TJ^T  $ and $\mu_i^{2D}=JW\mu_i$. The color $\mathbf{C}(u) $ of a pixel $u$ is rendered using the fast
$\alpha $-blending operation:
\begin{equation}
  \label{blending}
  \mathbf{C}(u) = \sum_{i\in N}T_i\alpha_i \mathbf{S} \mathbf{H}(h_i, v_i),
\end{equation}
where $T_i = \prod_{j=1}^{i-1}(1-\alpha_j)$, $\mathbf{S} \mathbf{H}$ is the spherical harmonic function and $v_i$ is the view direction.

%% file: method.tex
\subsection{Overview}
\label{overview}
\begin{figure}[!t]
    \vspace{-2em}
    \begin{center}
        \includegraphics[width=1\textwidth]{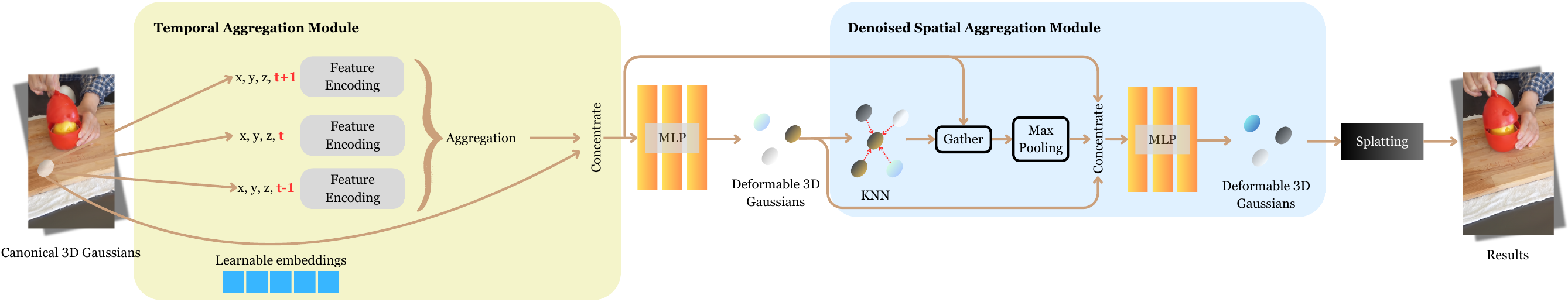}
        \caption{\textbf{The overall framework of our method DN-4DGS.} 
        Our approach employs a two-stage deformation process. In the first deformation, the well-designed Temporal Aggregation Module is utilized to aggregate temporal information. After the first deformation, the coordinate distribution of 3D gaussians is altered, and noise is suppressed. Subsequently, we proceed with the second deformation, utilizing the Denoised Spatial Aggregation Module to aggregate spatial information.
        }
        \label{framework}
    \end{center}
    \vspace{-0.5em}
\end{figure}

Our goal is to reconstruct dynamic 3D scenes from single/multi-view videos. Following previous works~\cite{wu20234d,yang2023deformable}, we represent the geometry and appearance of
the dynamic scene using canonical 3D gaussians and model the motion through the deformation fields. An overview of our method is shown in Figure~\ref{framework}. We first describe the details of the Noise Suppression Strategy (NSS) in Section~\ref{nss}. Then in Section~\ref{dts}, we present the design of the Decoupled Temporal-Spatial Aggregation Module (DTS). Section~\ref{o} details our optimization process.
\subsection{Noise Suppression Strategy}
\label{nss}
\begin{figure}[!t]
  \vspace{-2em}
  \begin{center}
      \includegraphics[width=1\textwidth]{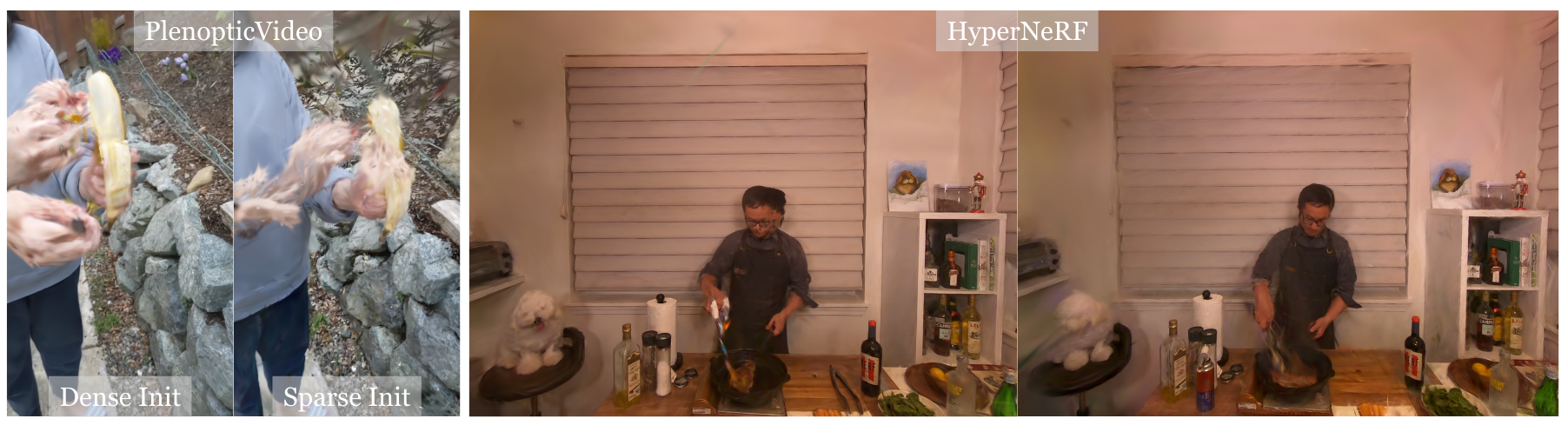}
      \caption{\textbf{More rendering images of canonical 3D gaussians.} 
      Here, $Sparse$ $Init$ refers to using the point cloud provided by the HyperNeRF~\cite{park2021hypernerf} dataset ($\mathrm{COLMAP_{SFM}}$~\cite{schonberger2016structure}) for Gaussian initialization, while $Dense$ $Init$ denotes generating a denser point cloud via $\mathrm{COLMAP_{MVS}}$~\cite{schonberger2016structure}. In fact, $Dense$ $Init$ can produce better rendering quality, but due to the need for regenerating, it consumes more computational resources.
      }
      \label{method_can}
  \end{center}
  \vspace{-1.3em}
\end{figure}

In this section, we attempt to mitigate the terrible noise of the canonical 3D gaussians, as shown in Figure~\ref{method_can}. 
Specifically, the Noise Suppression Strategy comprises two deformation operations.
The first deformation operation is a standard deformation. 
It takes the coordinates $x, y, z$ of the canonical 3D gaussians and time $t$ as input and outputs corresponding coordinate deformations $\Delta x, \Delta y, \Delta z$.
To simplify, here we only list the attributes $x, y, z$, while other attributes are detailed in the subsequent Section~\ref{dts},
\begin{equation}
  \label{d}
  \Delta x, \Delta y, \Delta z  = \Psi(x, y, z, T_n),
\end{equation}
where $\Psi$ represents the first deformation operation, $T_n$ represents the set of 
$t$'s neighbors. The details of $\Psi$ are introduced in Section~\ref{tag}. 
Next, after obtaining $\Delta x, \Delta y, \Delta z$, we add them to the original $x, y, z$.
\begin{equation}
  \label{update_xyz}
  x', y', z'  = x+\Delta x, y+\Delta y, z+\Delta z.
\end{equation}
Following this, the second deformation operation is carried out. 
\begin{equation}
  \label{second}
  \Delta x', \Delta y', \Delta z'  = \Psi'(P_k, t),
\end{equation}
where $\Psi'$ represents the second deformation operation, $P_k$ represents the set of $k$ neighbors of $(x', y', z')$. The details of $\Psi'$ are introduced in Section~\ref{sam}. 
In this deformation, due to the successful alteration of the input coordinate distribution during the first deformation stage, we can obtain more accurate Gaussian positions compared to the canonical Gaussian. As a result, the noise in the input is attenuated. 

Overall, Noise Suppression Strategy (NSS) is a strategy that uses two stages of deformation to reduce the impact of noise on the deformable network. During this process, we have two training phases. In the early training phase, we only supervise the first deformation. Once the Gaussian coordinates obtained from the first deformation are relatively accurate, we add the second deformation and shift the supervision to it.
\subsection{Decoupled Temporal-Spatial Aggregation Module}
\label{dts}
Local feature aggregation is very important for 3D point clouds, which can effectively extract local structure information. PointNet++~\cite{qi2017pointnet++} introduces the set abstraction layer to aggregate information from spatially adjacent points, which has become a fundamental operation in various point cloud tasks~\cite{pang2022masked,wu2022point,zhao2023divide,lu2023query,deng2024unsupervised}. Therefore, to enhance the accuracy of the deformation fields, an intuitive approach is to perform neighbor aggregation for each gaussian.

For 4D gaussians, there are four dimensions of information $x, y, z, t$ available for aggregation. As discussed in the introduction, performing local aggregation on noisy coordinates would further amplify the noise. 
Therefore, for $x, y, z$, spatial aggregation is conducted during the second deformation operation. Since temporal information is unrelated to the canonical 3D gaussians, temporal aggregation can be directly integrated into the first deformation operation to enhance feature extraction capabilities. 
To reduce computational overhead, and considering that temporal information has already been effectively extracted in the first deformation operation, we omit temporal aggregation in the second deformation operation.
The entire process is referred to as decoupled temporal-spatial aggregation.

\subsubsection{Temporal Aggregation Moudle}
\label{tag}
\begin{wrapfigure}[16]{r}{0.4\textwidth}
  \vspace{-2em}
  \begin{center}
      \includegraphics[width=0.8\textwidth]{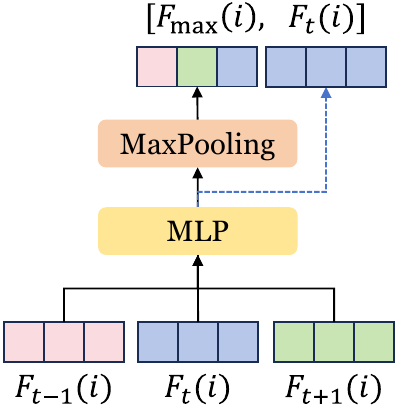}
      \caption{\textbf{The structure of aggregation operation.} 
      }
      \label{aggregation_op}
  \end{center}
\end{wrapfigure}

For each gaussian $G_t(i)$, we first input $x, y, z, t$ into the Feature Encoding. Regarding Feature Encoding, we can utilize the MLPs from D3DGS~\cite{yang2023deformable} or the HexPlanes from 4DGaussian~\cite{wu20234d}. After that, we acquire $F_t(i)$ at time $t$. Next is a critical step, where we repeat the above-mentioned process to obtain the features $F_{t-1}(i)$ for time $t-1$ and the features $F_{t+1}(i)$ for time $t+1$. It's worth noting that here, the "1" represents one timestep. Next, similar to PointNet++~\cite{qi2017pointnet++}, we perform the aggregation operation. As illustrated in Figure~\ref{aggregation_op}, we merge $F_{t-1}(i)$, $F_t(i)$, and $F_{t+1}(i)$ together to form $F(i)\in \mathbb{R}^{3\times 1\times C_1}$ and input into a lightweight MLP for channel change. 
Then, we perform MaxPooling along the first dimension of $F(i)$ to generate $F_{\text{max}}(i)\in \mathbb{R}^{ 1\times C_2}$. 
Additionally, we introduce a new attribute $\mathcal{Y}_i$, which is a learnable embedding. 
This attribute can provide information independent of coordinates without interference due to adjacency.  We have also observed a similar design in a recent work E-D3DGS~\cite{bae2024per}.
Finally, $F_{\text{max}}(i)$, $F_t(i)$ and $\mathcal{Y}_i$ are concatenated together to generate deformation: 
\begin{gather}
  \label{ffinal}
    F_t(i)' = [F_{\text{max}}(i), F_t(i), \mathcal{Y}_i], \\
    \mathcal{F}_\theta: F_t(i)'\rightarrow (\Delta x, \Delta y, \Delta z, \Delta r, \Delta s, \Delta \sigma ,\Delta h),
\end{gather}
where $\mathcal{F}_\theta$ is the deformation MLP head, $r$ is a rotation quaternion, $s$ is a vector for scaling, $\sigma$ is an opacity, and $h$ is
$\mathbf{S} \mathbf{H}$ coefficients for modeling view-dependent color.

\subsubsection{Denoised Spatial Aggregation Moudle}
\label{sam}
After obtaining the new $x', y', z'$, we input them into the denoised spatial aggregation module for spatial aggregation.
Concretely, we calculate the k-nearest neighbors for each gaussian based on \(x', y', z'\). Then, we aggregate the features of the k-nearest neighbors to obtain \(F_n(i) \in \mathbb{R}^{1 \times K \times C_3}\).
MaxPooling is then performed on $F_n(i)$ to get $F_{nm}(i)$. 
Finally, $F_n(i)$, $F_{nm}(i)$ and $x', y', z'$ are concatenated together to generate the second deformation: 
\begin{gather}
  \label{ffinal2}
    F_n(i)' = [F_n(i), F_{nm}(i), x', y', z'], \\
    \mathcal{F}_\theta': F_n(i)'\rightarrow (\Delta x, \Delta y, \Delta z, \Delta r, \Delta s, \Delta \sigma ,\Delta h),
\end{gather}
where $\mathcal{F}_\theta$ is the deformation MLP head in the second deformation operation.

Overall, Decoupled Temporal-Spatial Aggregation Module (DTS) is a specific feature aggregation method we propose. Unlike 4DGaussian, D3DGS lacks explicit spatiotemporal aggregation, so we designed DTS to aggregate spatiotemporal information. Considering that inaccurate coordinate relationships would be amplified through spatial aggregation (KNN), we only perform spatial aggregation in the second stage of NSS, which is DSAM. We name the spatial aggregation module "Denoised Spatial Aggregation Module" (DSAM) because the Gaussian coordinates input into DSAM are more accurate (denoised), as shown in the fourth column of Figure~\ref{two-stage}. Therefore, we prefix the Spatial Aggregation Module with "Denoised". DSAM itself does not have denoising capabilities; it solely performs spatial feature aggregation.
\subsection{Optimization}
\label{o}
The parameters to be optimized include the deformable network and the attributes of each 3D gaussian $G(i)$: $\mu_i$, $\sigma_i $, $\Sigma_i$, $h_i$ and $\mathcal{Y}_i$. Following 4DGaussian~\cite{wu20234d}, we use the reconstruction loss $\mathcal{L}_1$ and gird-based TV loss~\cite{hexplane,tineuvox,kplanes,dvgo} $\mathcal{L}_{tv}$ to supervise the training process. Additionally, we add a D-SSIM term $\mathcal{L}_{ssim}$ to improve structural similarity:
\begin{equation}
  \label{ffinal3}
  \mathcal{L}  = \lambda \mathcal{L}_1 + (1-\lambda )\mathcal{L}_{ssim} + \mathcal{L}_{tv},
\end{equation}
where $\lambda$ is the hyperparameter.
It is worth noting that we employ a two-stage training strategy. 
During the early stages of training, we exclusively execute the first deformation operation. Once satisfactory results are attained with the initial deformation operation, we then proceed to implement the second deformation operation to further refine accuracy.
The reason for this strategy is that only the deformation $\Delta x, \Delta y, \Delta z$ in the first stage is sufficiently precise to remove a large amount of noise, thereby positively impacting the deformation in the second stage.

%% file: expr.tex
\subsection{Experimental Setup}
\textbf{Dataset and Metrics.}
\textbf{PlenopticVideo}~\cite{li2022neural} dataset includes 20 multi-view videos, with each scene consisting of either 300 frames, except for the flame salmon scene, which comprises 1200 frames. These scenes encompass a relatively long duration and
various movements, with some featuring multiple objects in motion. We utilized PlenopticVideo dataset to observe the capability to capture dynamic
areas. Total six scenes (coffee martini, cook spinach, cut roasted beef, flame salmon, flame steak, sear steak) are utilized to train and render. 
Rendering resolution is set to 1352 $\times$ 1014
\textbf{HyperNeRF}~\cite{park2021hypernerf} dataset includes videos using two Pixel 3 phones rigidly mounted
on a handheld capture rig. We train and render on four scenes (3D Printer,
Banana, Broom, Chicken) at a resolution downsampled by a factor of two to
540 $\times$ 960.
\textbf{NeRF-DS}~\cite{yan2023nerf} dataset consists of seven captured videos (Sieve, Press, Plate, Cup, As, Bell, Basin) with camera pose
estimated using colmap~\cite{schonberger2016structure}. The dataset involves a variety of rigid and non-rigid deformation of various objects.
We train and render on the seven scenes. Rendering resolution is set to 480 $\times$ 270.

We report the quality of rendered images using PSNR, SSIM~\cite{ssim}, MS-SSIM and
LPIPS~\cite{lpips}. Higher PSNR, SSIM and MS-SSIM values and lower LPIPS
values indicate better visual quality. To PlenopticVideo dataset, we report PSNR, SSIM and LPIPS (Alex). To HyperNeRF dataset, we report PSNR, SSIM and MS-SSIM. 
To NeRF-DS dataset, we report PSNR, MS-SSIM and LPIPS (VGG).

\textbf{Implementation Details.}
\label{Implementation}
We train our model on a single RTX3090. The optimizer we utilize is Adam~\cite{diederik2014adam}. The learning rate is initially set at 1.6e-4, gradually decreasing exponentially to 1.6e-6 by the end of the training process. The learning rate for the voxel grid is initialized at 1.6e-3 and exponentially decays to 1.6e-5. 
For hyperparameters, we tune $K, \lambda $ as 16, 0.9 respectively. More details will be shown in the Appendix.

\subsection{Comparison with existing methods.}
\label{Comparisonwithexistingmethods}
\begin{figure}[!t]
  \begin{center}
      \includegraphics[width=0.95\textwidth]{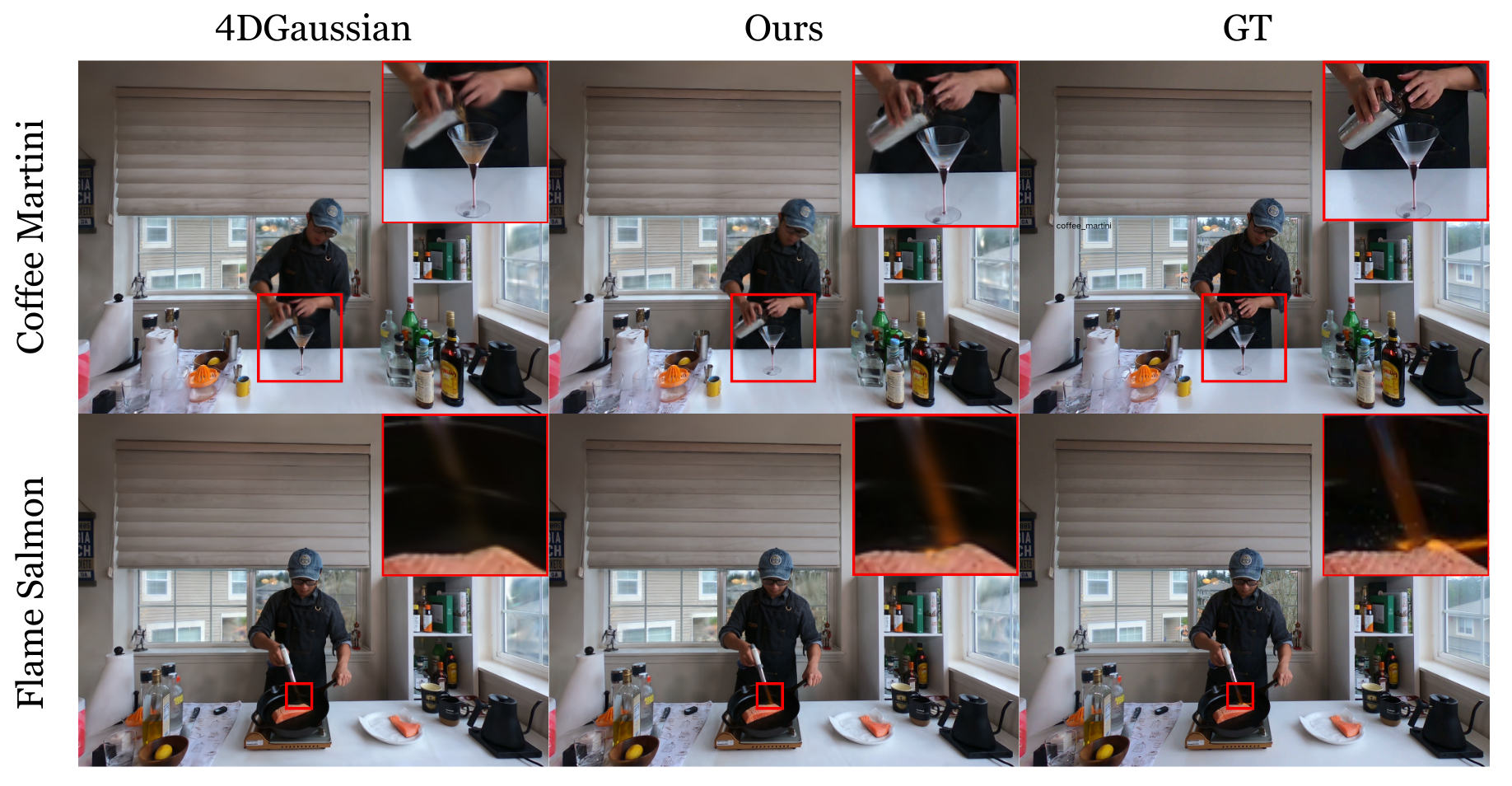}
      \caption{\textbf{Qualitative comparisons on PlenopticVideo Dataset.}}
      \label{dynerf}
  \end{center}
  \vspace{-1.3em}
\end{figure}

\textbf{Results on PlenopticVideo.}
Table~\ref{table:PlenopticVideo} reports the results on PlenopticVideo dataset. Refer to the Appendix for per-scene details.
Due to the incorporation of the Noise Suppression Strategy, which alters the distribution of canonical 3D gaussians coordinates to suppress noise, as well as the utilization of the Decoupled Temporal-Spatial Aggregation Module for feature aggregation, 
our approach demonstrates superior reconstruction quality across all metrics compared to the baseline (4DGaussian~\cite{wu20234d}). In fact, PSNR and LPIPS are currently the state-of-the-art metrics.
As the table shows, despite our method involving two stages of deformation, resulting in a slight weakening in training time and FPS, it overall meets the requirements for rapid training and real-time demands. 
Regarding storage, due to the presence of a new attribute $\mathcal{Y}_i$, it may slightly exceed the baseline.
To vividly illustrate the differences between our method and others, we visualize the qualitative results in Figure~\ref{dynerf}. From the regions highlighted in red boxes, it is evident that our method can render higher-quality images.
\begin{table}[!t]
  \begin{center}
    \footnotesize
    \setlength\tabcolsep{1.2pt}
    \caption{\textbf{Quantitative comparison on PlenopticVideo dataset.} 
    We display the average PSNR/SSIM/LPIPS (Alex) metrics for novel view synthesis on dynamic scenes, with each cell colored to indicate the \colorbox{yzybest}{best}, \colorbox{yzysecond}{second best}, and \colorbox{yzythird}{third best}. }
    \label{table:PlenopticVideo}
    \vspace{-0.13em}
    \begin{tabular}{c|ccc|c|cc}
      \toprule 
      Method & {\footnotesize{PSNR($\uparrow$)}} & {\footnotesize{SSIM($\uparrow$)}} & {\footnotesize{LPIPS($\downarrow$)}} & {\footnotesize{Time($\downarrow$)}} & {\footnotesize{FPS($\uparrow$)}} & {\footnotesize{Storage(MB)($\downarrow$)}} \\
      \midrule
      DyNeRF~\cite{li2022neural} & 29.58 &- &0.099&1344 hours &0.01 &\cellcolor{yzybest}28 \\
      NeRFPlayer~\cite{song2023nerfplayer} & 30.69 &0.909 &0.111 &6 hours&0.045&-\\
      HyperReel~\cite{attal2023hyperreel} & 31.10& 0.921 &0.096&9 hours&2.0&360\\
      HexPlane-all*~\cite{hexplane}& \cellcolor{yzysecond}31.70& \cellcolor{yzybest}0.984 &0.075&12 hours&0.2&250\\
      KPlanes~\cite{kplanes}& \cellcolor{yzythird}31.63&\cellcolor{yzysecond} 0.964 &-&\cellcolor{yzythird}1.8 hours&0.3&309\\
      \midrule
      4DGS~\cite{yang2023real}& 31.19 &0.940 &0.051 &9.5 hours &19.5& 8700\\
      E-D3DGS~\cite{bae2024per}& 31.31&\cellcolor{yzythird} 0.945 &\cellcolor{yzybest}0.037 &2 hours &\cellcolor{yzybest}43.1 &\cellcolor{yzysecond}35\\
      4DGaussian~\cite{wu20234d}& 31.15 &0.940 &\cellcolor{yzythird}0.049&\cellcolor{yzybest}40 mins&\cellcolor{yzysecond}30&\cellcolor{yzythird}90\\
      Ours& \cellcolor{yzybest}32.02 &0.944&\cellcolor{yzysecond}0.043&\cellcolor{yzysecond}50 mins &\cellcolor{yzythird}15&112\\
      \bottomrule
    \end{tabular}
   \vspace{-1.3em}
  \end{center}
\end{table}

\begin{figure}[!t]
  \begin{center}
      \includegraphics[width=0.95\textwidth]{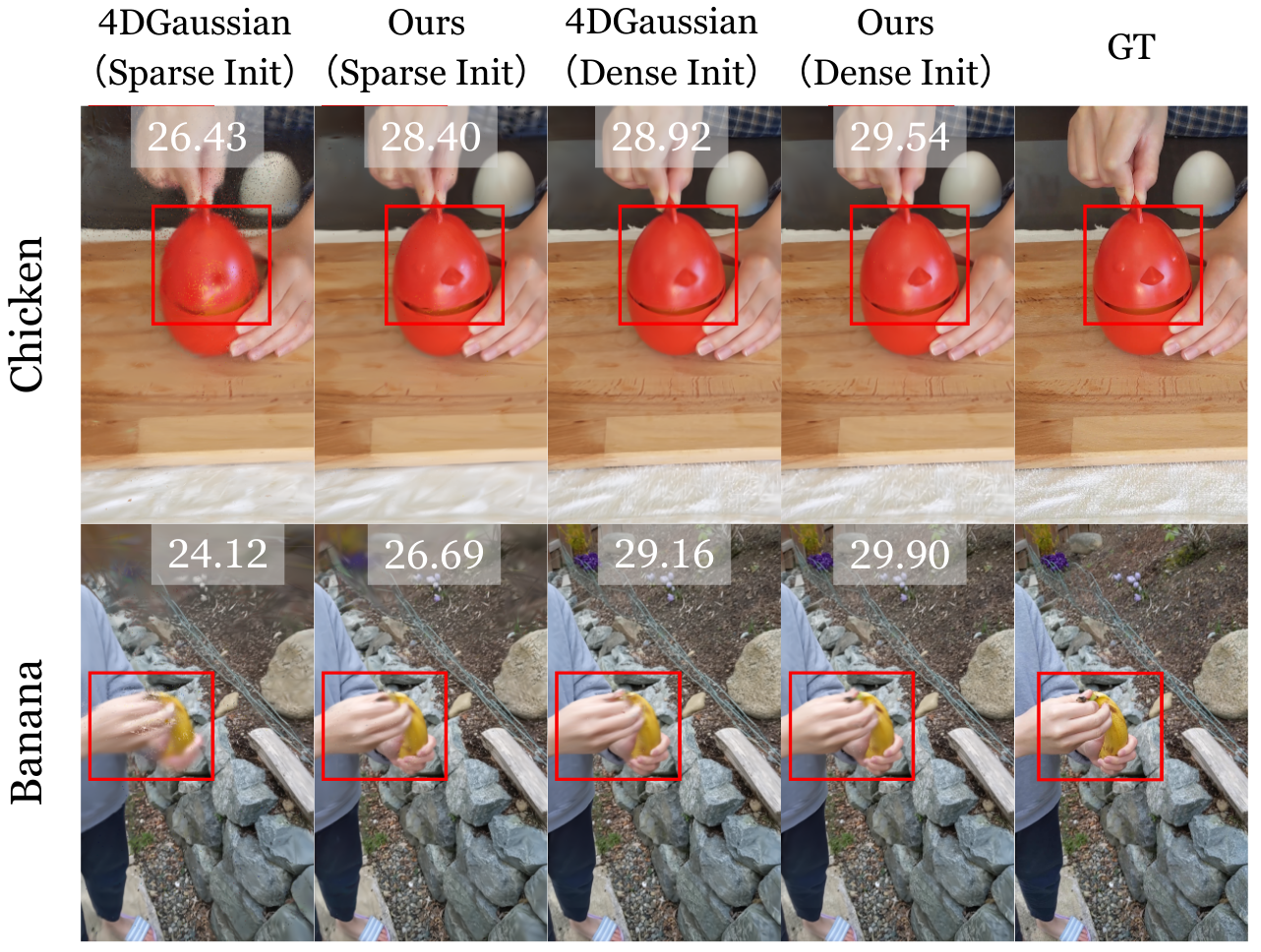}
      \caption{\textbf{Qualitative comparisons on HyperNeRF Dataset.} In the gray cells, the numbers represent PSNR.}
      \label{hypernerf1}
  \end{center}
  \vspace{-1.3em}
\end{figure}

\textbf{Results on HyperNeRF.}
Table~\ref{table:HyperNeRF} reports the results on HyperNeRF dataset. 
In this dataset, we present results for both $Sparse$ $Init$ and $Dense $ $Init$. 
$Sparse $ $Init$ refers to using the point cloud provided by the HyperNeRF dataset ($\mathrm{COLMAP_{SFM}}$~\cite{schonberger2016structure}) for Gaussian initialization, while $Dense $ $Init$ denotes generating a denser point cloud via $\mathrm{COLMAP_{MVS}}$~\cite{schonberger2016structure}.
From Table~\ref{table:HyperNeRF}, it can be observed that our method outperforms other Gaussian-based methods under both $Sparse $ $Init$ and $Dense $ $Init$ settings. 
Moreover, under the $Dense $ $Init$ setting, our method achieves the current state-of-the-art performance. 
More importantly, we find that 4DGaussian is highly sensitive to the sparsity or density of Gaussian initialization. 
In contrast, our approach benefits from noise suppression and feature aggregation, resulting in a more pronounced performance improvement under the sparse setting.
The qualitative results can be observed from Figure~\ref{hypernerf1}.
From the regions highlighted in red boxes, our method can render higher-quality images, as further supported by PSNR in gray cells.

\begin{table}[!t]
  \begin{center}
    \footnotesize
    \setlength\tabcolsep{1.2pt}
    \caption{\textbf{Quantitative comparison on HyperNeRF dataset.} Here, $\ddagger$ represents that we train the model based on $Sparse $ $Init$. * represents that we train the model based on $Dense $ $Init$.}
    \label{table:HyperNeRF}
    \vspace{-0.13em}
    \scalebox{0.95}{\begin{tabular}{c|ccc|c|cc}
      \toprule 
      Method & {\footnotesize{PSNR($\uparrow$)}} & {\footnotesize{SSIM($\uparrow$)}} & {\footnotesize{MS-SSIM($\uparrow$)}}& {\footnotesize{Time($\downarrow$)}} & {\footnotesize{FPS($\uparrow$)}} & {\footnotesize{Storage(MB)($\downarrow$)}} \\
      \midrule
      Nerfies~\cite{park2021nerfies} & 22.23 &-& 0.803&$\sim$ hours&<1&- \\
      HyperNeRF DS~\cite{park2021hypernerf} & 22.2 &0.598& 0.811&32 hours&<1&- \\
      TiNeuVox-B~\cite{tineuvox} & \cellcolor{yzythird}24.30 &0.616 &\cellcolor{yzythird}0.837 &\cellcolor{yzybest}30 mins&1&48 \\
      \midrule
      D3DGS$\ddagger$~\cite{yang2023deformable}&21.50 &-&-&2 hours&10&\cellcolor{yzythird}18\\
      4DGaussian$\ddagger$~\cite{wu20234d}&21.80&0.573&0.710&\cellcolor{yzysecond}50 mins&\cellcolor{yzybest}38&\cellcolor{yzybest}11\\
      Ours$\ddagger$ & 23.31 (\color{red}{+1.51}) & \cellcolor{yzythird}0.618 (\color{red}{+0.045})&0.768 (\color{red}{+0.058})&1.1 hours&\cellcolor{yzythird}24&\cellcolor{yzysecond}12\\
      \midrule
      D3DGS*&23.43 &-&-&3.5 hours&7&88\\
      4DGaussian*& \cellcolor{yzysecond}25.20 &\cellcolor{yzysecond}0.682 &\cellcolor{yzysecond}0.845&\cellcolor{yzythird}1 hour&\cellcolor{yzysecond}34&61\\
      Ours*& \cellcolor{yzybest}25.59 (\color{red}{+0.39})&\cellcolor{yzybest}0.691 (\color{red}{+0.009})&\cellcolor{yzybest}0.863 (\color{red}{+0.018})&1.2 hours&20&68\\
      \bottomrule
    \end{tabular}}
    \vspace{-1.3em}
  \end{center}
\end{table}

\textbf{Results on NeRF-DS.}
Table~\ref{table: NeRF-DS} presents the results on NeRF-DS dataset. 
Our proposed method achieves better performance compared to previous methods, demonstrating the effectiveness and generalization of our method. More qualitative results are shown in the Appendix.
\begin{figure}[!tp]
  \begin{floatrow}[2]
    \tablebox{\caption{\textbf{Quantitative comparison on D-NeRF dataset.} Here, parameter refers to the parameters of the deformation networks corresponding to different baselines.}\vspace{-1.3em}
    }{%
   \label{table:D-NeRF}
    \scalebox{0.65}{\begin{tabular}{c|ccc}
      \toprule 
      Method & {\footnotesize{PSNR($\uparrow$)}} & {\footnotesize{SSIM($\uparrow$)}} & {\footnotesize{Parameter (M)($\downarrow$)}} \\
      \midrule
      4DGaussian~\cite{wu20234d} & 34.05	&0.9787&3.38\\
      4DGaussian+Ours&34.53(\color{red}{+0.48})&	0.9811(\color{red}{+0.0024})&3.41(\color{blue}{+0.03})\\
      D3DGS~\cite{yang2023deformable}&	39.51&0.9902&0.52\\
      D3DGS+Ours&	39.87(\color{red}{+0.36})&0.9922(\color{red}{+0020})&0.39(\color{red}{-0.13})\\
      \bottomrule
    \end{tabular}}
    \vspace{-0.8em}
    }
  \figurebox{\caption{\textbf{The canonical results for both 4DGaussian and D3DGS.}}\vspace{-1.3em}}{%
  \label{figure:D-NeRF}
  \scalebox{0.5}{\includegraphics[width=0.95\textwidth]{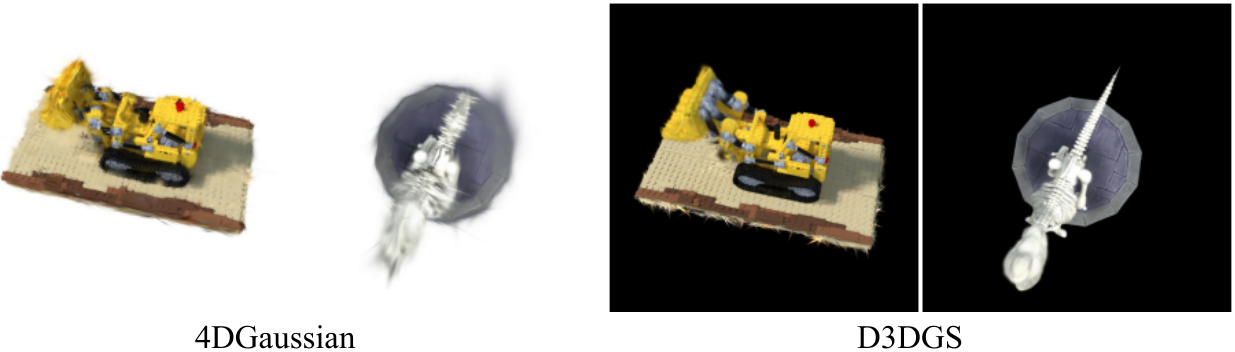}}
  \vspace{-0.8em}
  }
  
  \end{floatrow}
\end{figure}

\textbf{Results on D-NeRF.}
As illustrated in the Figure~\ref{figure:D-NeRF}, we have visualized the canonical results for both 4DGaussian and D3DGS. The results show that D3DGS has less noise compared to 4DGaussian, but noise is still present in moving areas. The quantitative results on D-NeRF dataset, as shown in the Table~\ref{table:D-NeRF}, indicate that our method improves performance on both 4DGaussian and D3DGS, with enhancements on 4DGaussian surpassing those on D3DGS. Regarding the parameters of the deformation networks, in 4DGaussian: Most of the parameters are composed of the HexPlane and deformation head. Our method introduces an additional deformation operation, resulting in a slight increase in the number of parameters compared to the baseline, with an increase of only 0.03M parameters. In D3DGS: The network primarily consists of an MLP-based deformation network. To reduce the computational load, our method halves the number of MLP layers, resulting in fewer overall parameters compared to the original D3DGS.
\subsection{Ablation Studies}

\textbf{Evaluation of the model with different designs.}
To evaluate the effectiveness of proposed components, we conduct an ablation study in Table~\ref{table:ablation} on PlenopticVideo dataset.
Here, NSS, TAM, DSAM represents the Noise Suppression Strategy, Temporal Aggregation Moudle and Denoised Spatial Aggregation Moudle respectively.
Specifically, the second row shows that with the use of two-stage deformation operations, our model can acquire a certain degree of improvement in quality.
This highlights the meaning of noise suppression.
The third row demonstrates that with the help of temporal aggregation, a performance gain of 0.41, 0.02, 0.003 has been achieved in PSNR, SSIM and LPIPS. 
The fourth row demonstrates the effective collaboration between NSS and TAM, resulting in performance improvement. 
The fifth row indicates that if we do not utilize two-stage training, but instead only perform spatial aggregation on canonical 3D gaussians, it not only fails to bring about an improvement in quality but also leads to a decrease.
The sixth row indicates that if we replace TAM with an ordinary deformation network, there is a slight drop in performance compared to the last row.
The last row indicates that if we combine all components together, the performance can reach its optimal level.
\begin{figure}[!tp]
  \begin{floatrow}[2]
    \tablebox{\caption{\textbf{Quantitative comparison on NeRF-DS dataset.}}\vspace{-0.8em}}{%
    \label{table: NeRF-DS}
    \scalebox{0.8}{
    \vspace{-0.13em}
    \begin{tabular}{c|ccc}
      \toprule 
      Method & {\footnotesize{PSNR($\uparrow$)}} & {\footnotesize{SSIM($\uparrow$)}} & {\footnotesize{LPIPS($\downarrow$)}} \\
      \midrule
      TiNeuVox~\cite{tineuvox} &  21.61 &0.823 &0.277 \\
      HyperNeRF~\cite{park2021hypernerf} & 23.45 &\cellcolor{yzythird}0.849 &0.199\\
      NeRF-DS~\cite{yan2023nerf} & \cellcolor{yzythird}23.60 &\cellcolor{yzythird}0.849 &\cellcolor{yzythird}0.182\\
      \midrule
      3D-GS~\cite{3dgs} &  20.29 &0.782 &0.292\\
      D3DGS~\cite{yang2023deformable}& \cellcolor{yzysecond}23.92 &\cellcolor{yzysecond}0.847 &\cellcolor{yzysecond}0.184\\
      Ours& \cellcolor{yzybest}24.36 &\cellcolor{yzybest}0.865&\cellcolor{yzybest}0.171\\
      \bottomrule
    \end{tabular}}
    \vspace{-0.8em}
    }
  \tablebox{\caption{\textbf{Evaluation of the model with different designs on PlenopticVideo dataset.} }\vspace{-0.8em}}{%
  \label{table:ablation}
  \scalebox{0.8}{\begin{tabular}{ccc|ccc}
      \toprule
      NSS & TAM & DSAM  &{\footnotesize{PSNR($\uparrow$)}} & {\footnotesize{SSIM($\uparrow$)}} & {\footnotesize{LPIPS($\downarrow$)}} \\
      \midrule
      \ding{55}&\ding{55}&\ding{55}& 31.15 & 0.940  & 0.049  \\
      \ding{51}&\ding{55}&\ding{55}& 31.31&0.941&0.047	\\
      \ding{55}&\ding{51}&\ding{55}&31.56&0.942&0.046\\
      \ding{51}&\ding{51}&\ding{55}&31.69&0.943&0.045\\
      \ding{55}&\ding{51}&\ding{51}&31.31&0.936 &0.052\\
      \ding{51}&\ding{55}&\ding{51}&31.72&0.943&0.045\\
      \ding{51}&\ding{51}&\ding{51}	&\textbf{32.02}&\textbf{0.944} &\textbf{0.043}
      \\
      \bottomrule
  \end{tabular}}
  \vspace{-0.8em}
  }
  
  \end{floatrow}
\end{figure}

\begin{figure}[!t]
  \begin{center}
      \includegraphics[width=0.95\textwidth]{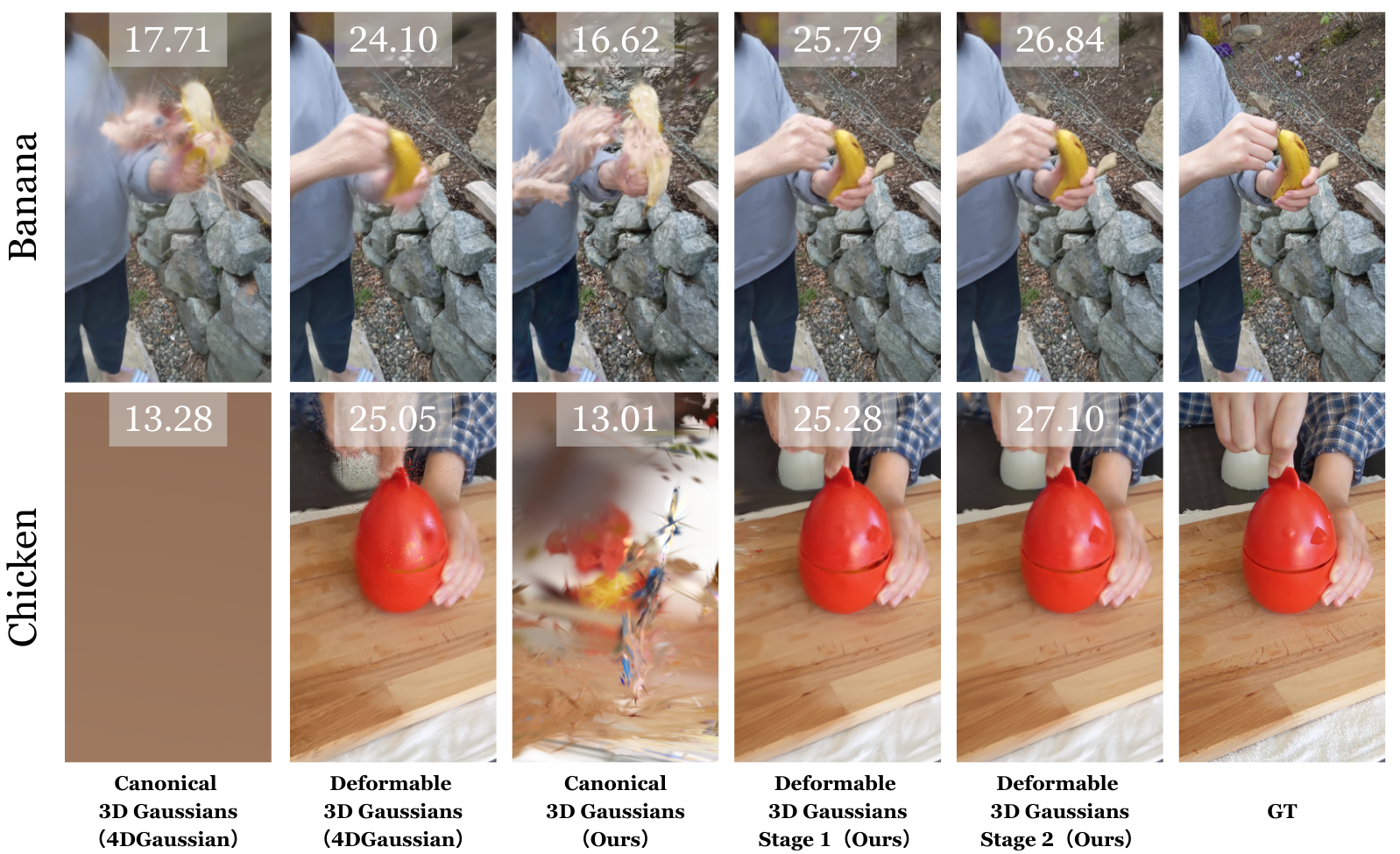}
      \caption{\textbf{Effectiveness of the two-stage deformation operations.} In the gray cells, the numbers represent
PSNR.\vspace{-1.3em}}
      \label{two-stage}
  \end{center}
  \vspace{-1.3em}
\end{figure}

\textbf{Effectiveness of the two-stage deformation operations.}
To validate the significance of two-stage deformation operations, we conducted visual experiments. As shown in Figure~\ref{two-stage}, canonical 3D aussians exhibit a significant amount of noise, which severely affects the accuracy of the deformation field.
To alleviate this issue, we employed two-stage deformation. 
As depicted in the fourth column of the figure, after the first deformation, there is a significant change in the distribution of coordinates $xyz$, effectively suppressing the noise. 
Moreover, due to the design of temporal aggregation, the corresponding PSNR value is even higher than that of 4DGaussian.
Finally, by performing the second deformation operation on the basis of the first-stage deformation, the performance is further improved. 
This is attributed to the cleaner coordinates $xyz$ and the design of spatial aggregation.

\subsection{Limitations and Future Work}
\label{Limitations}
Although two-stage deformation can alter the coordinate distribution of canonical 3D gaussians and reduce the noise introduced into the deformation field, the lack of simultaneous supervision for both stages~\cite{carion2020end,cheng2022masked,lu2023query} poses a challenge. 
Consequently, during the second stage, due to the lack of supervision in the first-stage deformation, the direction of coordinate deformation becomes uncontrollable to some extent. 
This, in turn, affects the spatial feature aggregation in the second stage.
To address this issue, future work should explore the direction of simultaneous supervision.

%% file: clu.tex
In this paper, we introduce a novel representation called Denoised Deformable Network with Temporal-Spatial Aggregation for Dynamic Scene Rendering.
We promose the Noise Suppression Strategy, which can change the distribution of the coordinates of the canonical 3D gaussians, suppress noise and generate a more precise deformation field.
To aggregate information from adjacent points and frames, we promose the Decoupled Temporal-Spatial Aggregation Module.
Extensive experiments on various real-world datasets demonstrate that our method achieves state-of-the-art rendering quality under a real-time level.

%% file: supplementary.tex
\subsection{Overview}

This supplementary material provides more model and experimental details to understand our proposed method. After that, we present more experiments to demonstrate the effectiveness of our methods. Finally, we show a rich visualization of our modules. 

\subsection{More Model Details}
\textbf{Feature Encoding.}  
As illustrated in Section~\ref{tag}, we can utilize the MLPs from D3DGS~\cite{yang2023deformable} or the HexPlanes from 4DGaussian~\cite{wu20234d} as Feature Encoding.
Specifically, for both PlenopticVideo~\cite{li2022neural} and HyperNeRF~\cite{park2021hypernerf}, we use the HexPlanes to encode per-gaussian's feature. The complete details can be referred to 4DGaussian's main text.
For NeRF-DS~\cite{yan2023nerf}, we utilize the MLPs for feature encoding. The complete details can be referred to D3DGS's main text. 

\subsection{More Implementation Details}
For both PlenopticVideo and HyperNeRF, the total training comprises 14,000 iterations, with the first stage encompassing 5,000 iterations.
For NeRF-DS, the total training comprises 40,000 iterations, with the first stage encompassing 15,000 iterations.
The dimension of attribute $\mathcal{Y}$ is set as 16.

\subsection{Comparison with 3DGStream}
\textbf{Similarities:} Our method and 3DGStream both perform deformations on 3D Gaussians (3DGs) where absolute positions and relative positional relationships are more accurately maintained. For each timestep $i$, 3DGStream uses the 3DGs from the previous timestep $i-1$ as initialization. In contrast, our method uses canonical Gaussians (which are time-independent) as the initialization. To achieve accurate relative positional relationships and minimize noise interference from the canonical Gaussians, we employ a two-stage deformation strategy. The first stage obtains accurate 3DGs, and the second stage further deforms these 3DGs to achieve preciser rendering results.

\textbf{Advantages:}
1. \textit{Flexibility:} Unlike 3DGStream, our method does not require the results from the previous timestep. We can render at any arbitrary time without needing the previous timestep's 3DGs. On the other hand, 3DGStream relies on the 3DGs from the previous timestep for rendering the next.
2. \textit{Robustness:} 3DGStream heavily depends on the reconstruction quality at timestep 0. If the initial reconstruction is poor, subsequent reconstructions will be negatively affected, and these errors can accumulate over time. Our method, however, starts with noisy canonical Gaussians and improves the reconstruction quality through a two-stage deformation process, resulting in progressively better reconstructions.

\textbf{Disadvantages:}
\textit{Training Efficiency:} 3DGStream employs an online reconstruction approach, leading to shorter training time and faster rendering speed. In contrast, our method involves offline training, which results in relatively longer training time.
\subsection{Detailed Results}
In Table~\ref{tab:PlenopticVideo_comparison}, Table~\ref{tab:hypernerf_comparison} and Table~\ref{table:nerf-ds}, we provide the results for individual scenes associated with Section~\ref{Comparisonwithexistingmethods} of the main text. It can be observed that
our method achieved superior metrics in almost every scene compared to previous methods, demonstrating the effectiveness and generalization of our method under various scenes.
\begin{table}
  \centering  
  \begin{threeparttable}  
  \caption{\textbf{Per-scene results of PlenopticVideo dataset.}}  
  \label{tab:PlenopticVideo_comparison}  
  \begin{tabular}{c cc cc cc}  
      \toprule  
      \multirow{2}{*}{Method}&  
      \multicolumn{2}{c}{Cut Beef}&\multicolumn{2}{c}{Cook Spinach}&\multicolumn{2}{c}{Sear Steak}\cr  
      \cmidrule(lr){2-3}\cmidrule(lr){4-5}\cmidrule(lr){6-7}
      &PSNR&SSIM&PSNR&SSIM&PSNR&SSIM\cr  
      \midrule  
      NeRFPlayer~\cite{song2023nerfplayer}&31.83&0.928&32.06&0.930&32.31&0.940\cr
      HexPlane~\cite{hexplane}&\cellcolor{yzythird}32.71&\cellcolor{yzybest}0.985&31.86&\cellcolor{yzybest}0.983&32.09&\cellcolor{yzybest}0.986\cr
      KPlanes~\cite{kplanes}&31.82&\cellcolor{yzysecond}0.966&\cellcolor{yzysecond}32.60&\cellcolor{yzysecond}0.966&\cellcolor{yzysecond}32.52&\cellcolor{yzysecond}0.974\cr
      MixVoxels~\cite{wang2023mixedvoxels}&31.30&\cellcolor{yzythird}0.965&31.65&\cellcolor{yzythird}0.965&31.43&\cellcolor{yzythird}0.971\cr
      4DGaussian~\cite{wu20234d}&\cellcolor{yzysecond}32.90&0.957&\cellcolor{yzythird}32.46&0.949&\cellcolor{yzythird}32.49&0.957\cr
      Ours&\cellcolor{yzybest}33.49&0.960&\cellcolor{yzybest}32.91&0.951&\cellcolor{yzybest}33.98&0.959\cr
  \toprule  
      \multirow{2}{*}{Method}&  
      \multicolumn{2}{c}{Flame Steak}&\multicolumn{2}{c}{Flame Salmon}&\multicolumn{2}{c}{Coffee Martini}\cr  
      \cmidrule(lr){2-3}\cmidrule(lr){4-5}\cmidrule(lr){6-7}
      &PSNR&SSIM&PSNR&SSIM&PSNR&SSIM\cr  
      \midrule
      NeRFPlayer~\cite{song2023nerfplayer}&27.36&0.867&26.14&0.849&\cellcolor{yzybest}32.05&\cellcolor{yzythird}0.938\cr
      HexPlane~\cite{hexplane}&31.92&\cellcolor{yzybest}0.988&\cellcolor{yzythird}29.26&\cellcolor{yzybest}0.980&-&-\cr
      KPlanes~\cite{kplanes}&\cellcolor{yzythird}32.39&\cellcolor{yzysecond}0.970&\cellcolor{yzybest}30.44&\cellcolor{yzysecond}0.953&\cellcolor{yzysecond}29.99&\cellcolor{yzybest}0.953\cr
      MixVoxels~\cite{wang2023mixedvoxels}&31.21&\cellcolor{yzysecond}0.970&\cellcolor{yzysecond}29.92&\cellcolor{yzythird}0.945&\cellcolor{yzythird}29.36&\cellcolor{yzysecond}0.946\cr
      4DGaussian~\cite{wu20234d}&\cellcolor{yzysecond}32.51&0.954&29.20&0.917&27.34&0.905\cr
      Ours&\cellcolor{yzybest}33.51&\cellcolor{yzythird}0.958&29.19&0.921&29.04&0.915\cr
      \bottomrule  
  \end{tabular}  
  \end{threeparttable}  
  \end{table}  
  
  \begin{table}	
    \centering  
    \begin{threeparttable}  
    \caption{\textbf{Perscene results of HyperNeRF dataset by different models.} Here, $\ddagger$ represents that we train the model based on $Sparse $ $Init$. * represents that we train the model based on $Dense $ $Init$.}  
    \label{tab:hypernerf_comparison}  
    \scalebox{0.85}{\begin{tabular}{c cc cc cc cc}  
    \toprule  
    \multirow{2}{*}{Method}&  
    \multicolumn{2}{c}{3D Printer}&\multicolumn{2}{c}{Chicken}&\multicolumn{2}{c}{Broom}&\multicolumn{2}{c}{Banana}\cr  
    \cmidrule(lr){2-3}\cmidrule(lr){4-5}\cmidrule(lr){6-7}\cmidrule(lr){8-9}
    &PSNR&MS-SSIM&PSNR&MS-SSIM&PSNR&MS-SSIM&PSNR&MS-SSIM\cr  
    \midrule  
    Nerfies~\cite{park2021nerfies}&20.6&\cellcolor{yzythird}0.83&26.7&0.94&19.2&0.56&22.4&0.87\cr  HyperNeRF~\cite{park2021hypernerf}&20.0&0.59&26.9&0.94&19.3&0.59&23.3&0.90\cr  
    TiNeuVox-B~\cite{tineuvox}&\cellcolor{yzybest}22.8&\cellcolor{yzybest}0.84&\cellcolor{yzythird}28.3&\cellcolor{yzysecond}0.95&21.5&0.69&24.4&\cellcolor{yzythird}0.87\cr  
    FFDNeRF~\cite{guo2023forwardflowfornvs}&\cellcolor{yzysecond}22.8&\cellcolor{yzysecond}0.84&28.0&\cellcolor{yzythird}0.94&\cellcolor{yzythird}21.9&0.71\cellcolor{yzysecond}&24.3&0.86\cr
    3D-GS~\cite{3dgs}&18.3&0.60&19.7&0.70&20.6&0.63&20.4&0.80\cr
    4DGaussian~\cite{wu20234d}$\ddagger$ &20.9&0.75&24.1&0.82&20.0&0.53&22.2&0.74\cr
    Ours$\ddagger$ &21.6&0.78&26.2&0.89&20.8&0.57&\cellcolor{yzythird}24.7&0.84\cr
    4DGaussian* &22.1&0.81&\cellcolor{yzysecond}28.7&0.93&\cellcolor{yzysecond}22.0&0.70\cellcolor{yzythird}&\cellcolor{yzysecond}28.0&\cellcolor{yzysecond}0.94\cr
    Ours* &\cellcolor{yzythird}22.1&0.81&\cellcolor{yzybest}29.2&\cellcolor{yzybest}0.95&\cellcolor{yzybest}22.3&\cellcolor{yzybest}0.74&\cellcolor{yzybest}28.7&\cellcolor{yzybest}0.95\cr
    \bottomrule  
    \end{tabular} } 
    \end{threeparttable}  
  \end{table}  
    
  \begin{table}[!t]
    \begin{center}
      \footnotesize
      \setlength\tabcolsep{3pt}
      \caption{\textbf{Perscene results of NeRF-DS dataset by different models.} }
      \label{table:nerf-ds}
      \vspace{-0.13em}
      \scalebox{1}{\begin{tabular}{c|ccccccc}
        \toprule 
        \multirow{2}{*}{Method}&  \multicolumn{7}{c}{PSNR} \\
        \cmidrule{2-8}
        & Sieve &  Plate& Bell&Press&Cup&As&Basin\\
        \midrule
        3D-GS~\cite{3dgs} &23.16 &16.14 & 21.01 & 22.89 & 21.71 &22.69 &18.42\\
        TiNeuVox~\cite{tineuvox}&  21.49& \cellcolor{yzysecond}20.58  &23.08& 24.47&  19.71&  21.26 &\cellcolor{yzybest}20.66 \\
        HyperNeRF~\cite{park2021hypernerf} & 25.43 &18.93& 23.06& \cellcolor{yzysecond}26.15& \cellcolor{yzythird}24.59 &\cellcolor{yzythird}25.58 &\cellcolor{yzysecond}20.41\\
        NeRF-DS~\cite{yan2023nerf} &\cellcolor{yzysecond}25.78& \cellcolor{yzythird}20.54 &\cellcolor{yzythird}23.19&\cellcolor{yzythird} 25.72 &\cellcolor{yzysecond}24.91& 25.13 &\cellcolor{yzythird}19.96\\
        D3DGS~\cite{yang2023deformable} & \cellcolor{yzythird}25.72& 20.40 &\cellcolor{yzysecond}25.24 &25.70& 24.35& \cellcolor{yzysecond}26.35 &19.70 \\
        Ours & \cellcolor{yzybest}26.60 & \cellcolor{yzybest}20.92& \cellcolor{yzybest}25.48& \cellcolor{yzybest}26.24& \cellcolor{yzybest}24.95& \cellcolor{yzybest}26.54& 19.79\\
        \bottomrule
      \end{tabular}}
    \end{center}
  \end{table}

\subsection{More Ablation Studies}

\textbf{The effectiveness of $\mathcal{Y}$.} As depicted in Table~\ref{table:mathcal{Y}}, we conduct an ablation study on the flame steak scene of PlenopticVideo dataset to examine the impact of $\mathcal{Y}$. 
From the table, it is evident that $\mathcal{Y}$ can enhance rendering quality and $\mathcal{Y}$=16 is the best setting.
\begin{table}[!t]
  \begin{center}
    \footnotesize
    \setlength\tabcolsep{5pt}
    \centering
    \caption{\textbf{The effectiveness of $\mathcal{Y}$.} We compare the results on the flame steak scene of PlenopticVideo dataset.}
    \label{table:mathcal{Y}}
    \begin{tabular}{c|c|cc}
      \toprule
      Setting &Dim & PSNR & SSIM \\
      \midrule
      W/o $\mathcal{Y}$ &-&33.13&0.956\\
      W $\mathcal{Y}$ &4&33.33&0.957\\
      W $\mathcal{Y}$ &16 &\textbf{33.51} &\textbf{0.958}\\
      W $\mathcal{Y}$ &32&33.40&0.958\\
      W $\mathcal{Y}$ &64&33.29&0.957\\
      \bottomrule 
    \end{tabular}
  \end{center}
\end{table}

\textbf{Ablation study of timestep in Temporal Aggregation Moudle.} 
From Table~\ref{table:timestep}, it is evident that when timestep is set as 1$\times$, the performance is the best.
We speculate that the reason for the lowest performance at 0.5$\times$ might be attributed to the absence of images corresponding to $\pm$0.5 timestep in the dataset. Consequently, the lack of supervision at this timestep may induce significant feature biases, resulting in relatively poorer performance.
\begin{table}[!t]
  \begin{center}
    \footnotesize
    \setlength\tabcolsep{5pt}
    \centering
    \caption{\textbf{Ablation study of timestep in Temporal Aggregation Moudle.} We compare the results on the flame steak scene of PlenopticVideo dataset.}
    \label{table:timestep}
    \begin{tabular}{c|cc}
      \toprule
      Setting  & PSNR & SSIM \\
      \midrule
      0.5$\times$ &33.40 & 0.957 \\
      1$\times$ & \textbf{33.51} &\textbf{0.958}\\
      2$\times$ &33.45& 0.957 \\
      \bottomrule 
    \end{tabular}
  \end{center}
\end{table}

\textbf{Ablation study of  $K$ in Denoised Spatial Aggregation Moudle.} 
From Table~\ref{table:K}, it is evident that setting $ K = 16 $ yields the best results. 
\begin{table}[!t]
  \begin{center}
    \footnotesize
    \setlength\tabcolsep{5pt}
    \centering
    \caption{\textbf{Ablation study of $K$ in Denoised Spatial Aggregation Moudle.} We compare the results on the flame steak scene of PlenopticVideo dataset.}
    \label{table:K}
    \begin{tabular}{c|cc}
      \toprule
      $K$  & PSNR & SSIM \\
      \midrule
      4 & 33.39 & 0.957\\
      16 & \textbf{33.51} &\textbf{0.958}\\
      32 &33.35 &0.957\\
      \bottomrule 
    \end{tabular}
  \end{center}
\end{table}

\textbf{Ablation study of the iteration rounds of the first stage.}
From Table~\ref{table:iteration}, it is evident that the two-stage training strategy is meaningful.
If we train both deformation operations simultaneously, the performance is poor, as indicated in the first row. 
Only by first training the first deformation network and then proceeding to train the second deformation network after the first deformation network is well-trained can we achieve optimal performance. Additionally, the optimal number of iteration roudns is 8000.
\begin{table}[!t]
  \begin{center}
    \footnotesize
    \setlength\tabcolsep{5pt}
    \centering
    \caption{\textbf{Ablation study of the iteration rounds of the first stage.} We compare the results on the flame steak scene of PlenopticVideo dataset.}
    \label{table:iteration}
    \begin{tabular}{c|ccc}
      \toprule
      Rounds  & PSNR & SSIM \\
      \midrule
      0 & 32.89&0.951 \\
      4000 &33.40&0.957 \\
      6000 &33.38 & 0.957 \\
      8000 &\textbf{33.51} &\textbf{0.958}\\
      10000&33.37 &0.957\\
      \bottomrule 
    \end{tabular}
  \end{center}
\end{table}

\textbf{Ablation study of the iteration rounds of the first stage.}

\begin{figure}[!t]
  \begin{center}
      \includegraphics[width=0.95\textwidth]{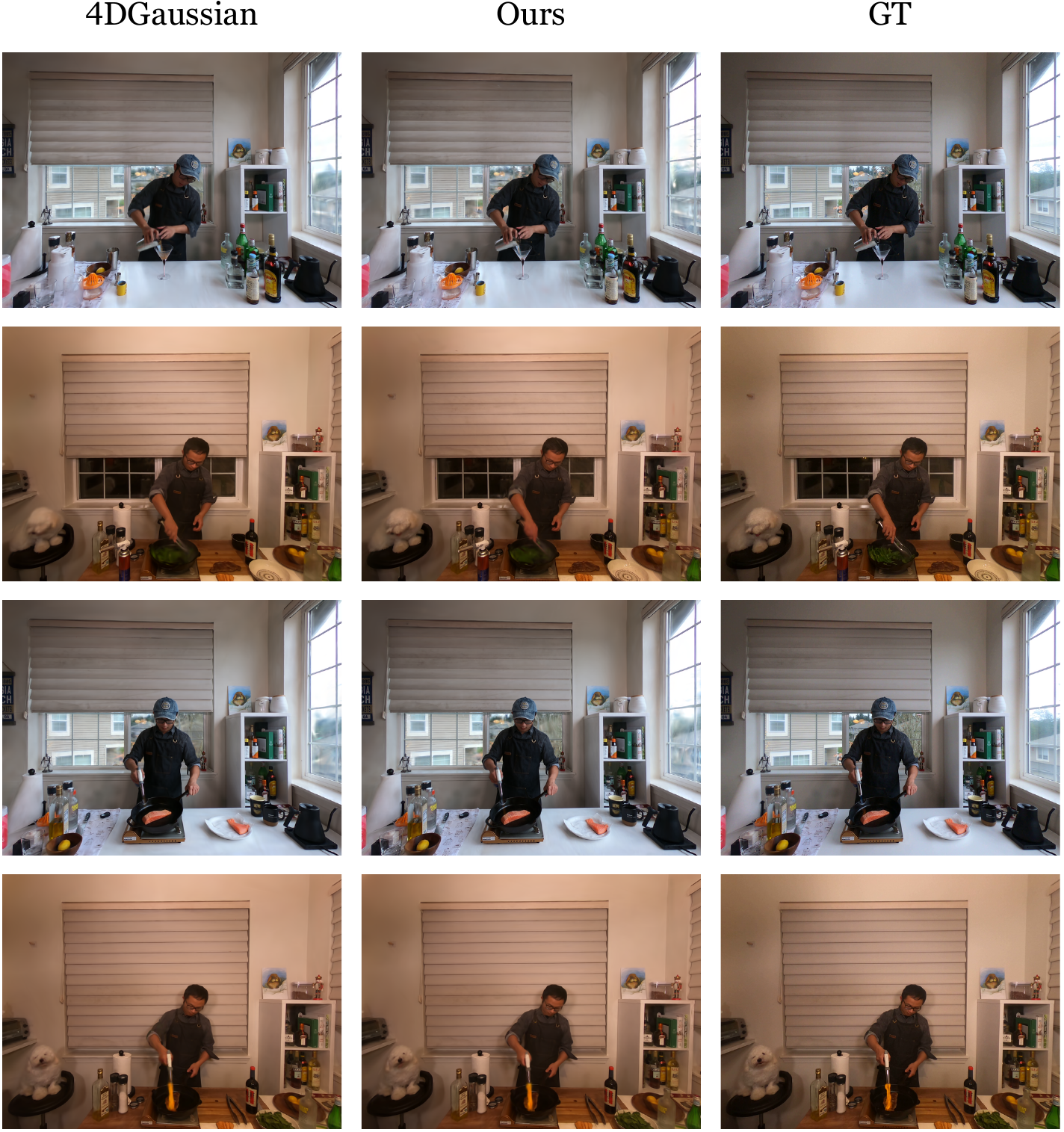}
      \caption{\textbf{Qualitative comparisons on PlenopticVideo Dataset.}}
      \label{PlenopticVideo}
  \end{center}
  \vspace{-1.3em}
\end{figure}
\subsection{Assets Availability}
\label{asset}
The datasets that support the findings of this study are available in the following repositories: 
HyperNeRF~\cite{park2021hypernerf} at \url{https://github.com/google/hypernerf/releases/tag/v0.1} under Apache-2.0 license.
NeRF-DS~\cite{yan2023nerf} at \url{https://github.com/JokerYan/NeRF-DS/releases/tag/v0.1-pre-release} under Apache-2.0 license.
PlenopticVideo~\cite{li2022neural} at \url{https://github.com/facebookresearch/Neural_3D_Video?tab=License-1-ov-file} under CC BY-NC 4.0 license.
The code of our baseline~\cite{wu20234d,yang2023deformable} is available at \url{https://github.com/ingra14m/Deformable-3D-Gaussians} under MIT license and \url{https://github.com/hustvl/4DGaussians} under Gaussian-Splatting license.

\subsection{More Visual Comparison}
Figure~\ref{PlenopticVideo} shows more visual comparisons on PlenopticVideo Dataset. 
We compare the results of 4DGaussian and our model. 

\begin{figure}[!ht]
  \begin{center}
      \includegraphics[width=0.95\textwidth]{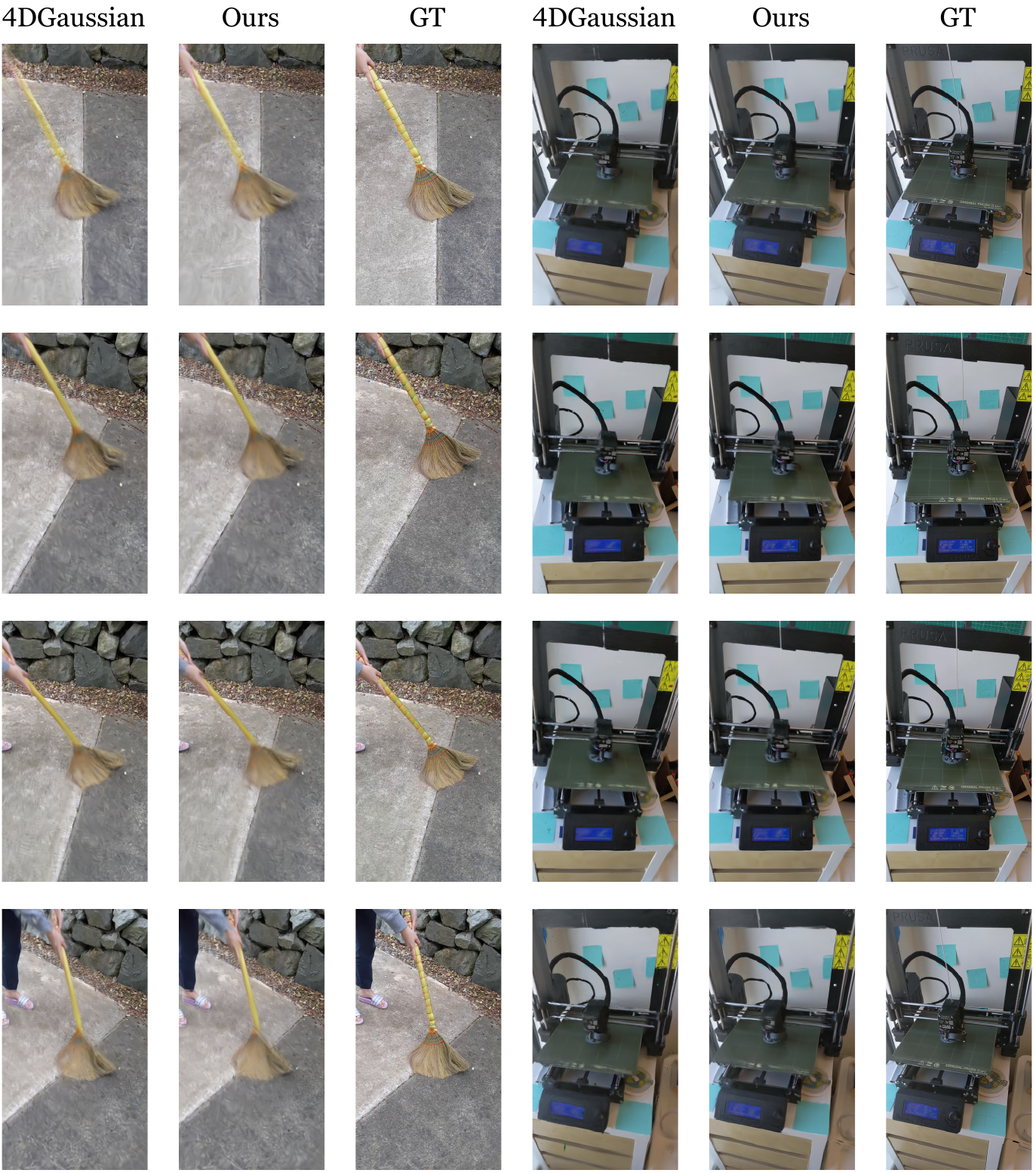}
      \caption{\textbf{Qualitative comparisons on HyperNeRF Dataset.}}
      \label{HyperNeRF}
  \end{center}
  \vspace{-1.3em}
\end{figure}

\begin{figure}[!ht]
  \begin{center}
      \includegraphics[width=0.95\textwidth]{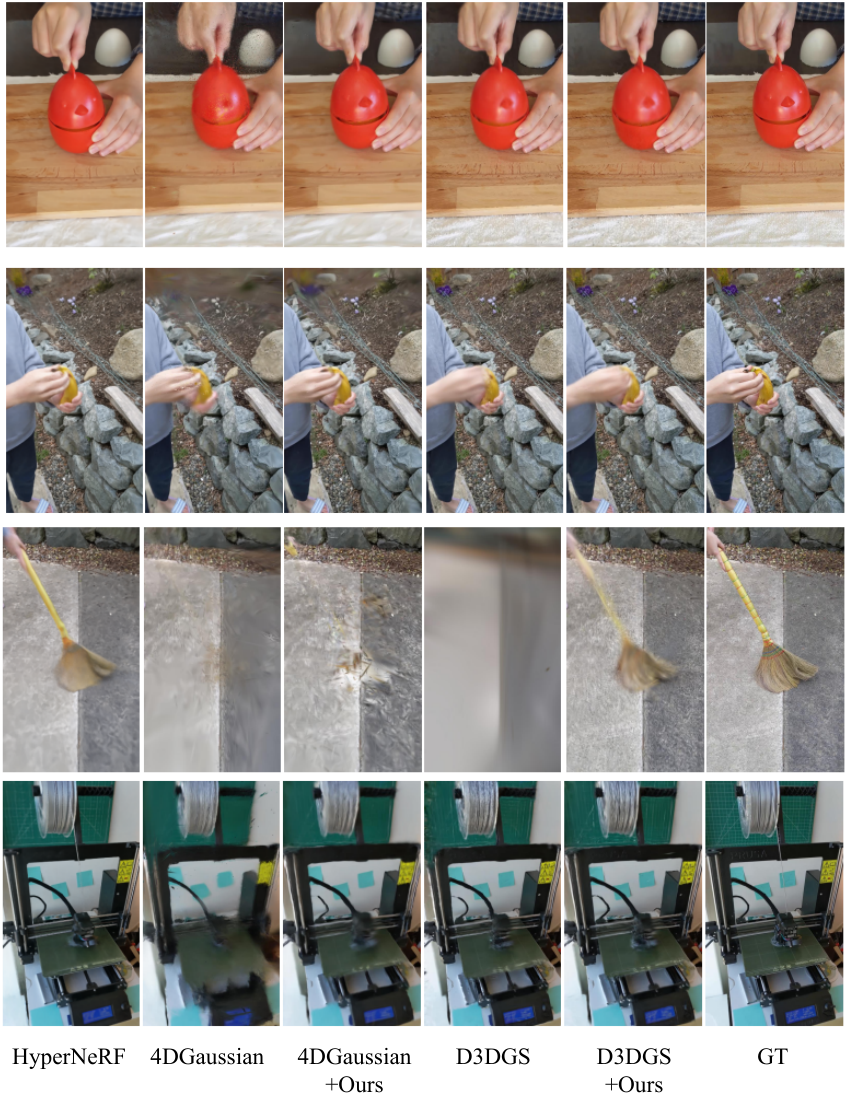}
      \caption{\textbf{Qualitative comparisons on HyperNeRF Dataset.}}
      \label{HyperNeRF1}
  \end{center}
  \vspace{-1.3em}
\end{figure}

Figure~\ref{HyperNeRF} and Figure~\ref{HyperNeRF1} shows more visual comparisons on HyperNeRF Dataset. 
We compare the results of 4DGaussian and our model.

\begin{figure}[!ht]
  \begin{center}
      \includegraphics[width=0.95\textwidth]{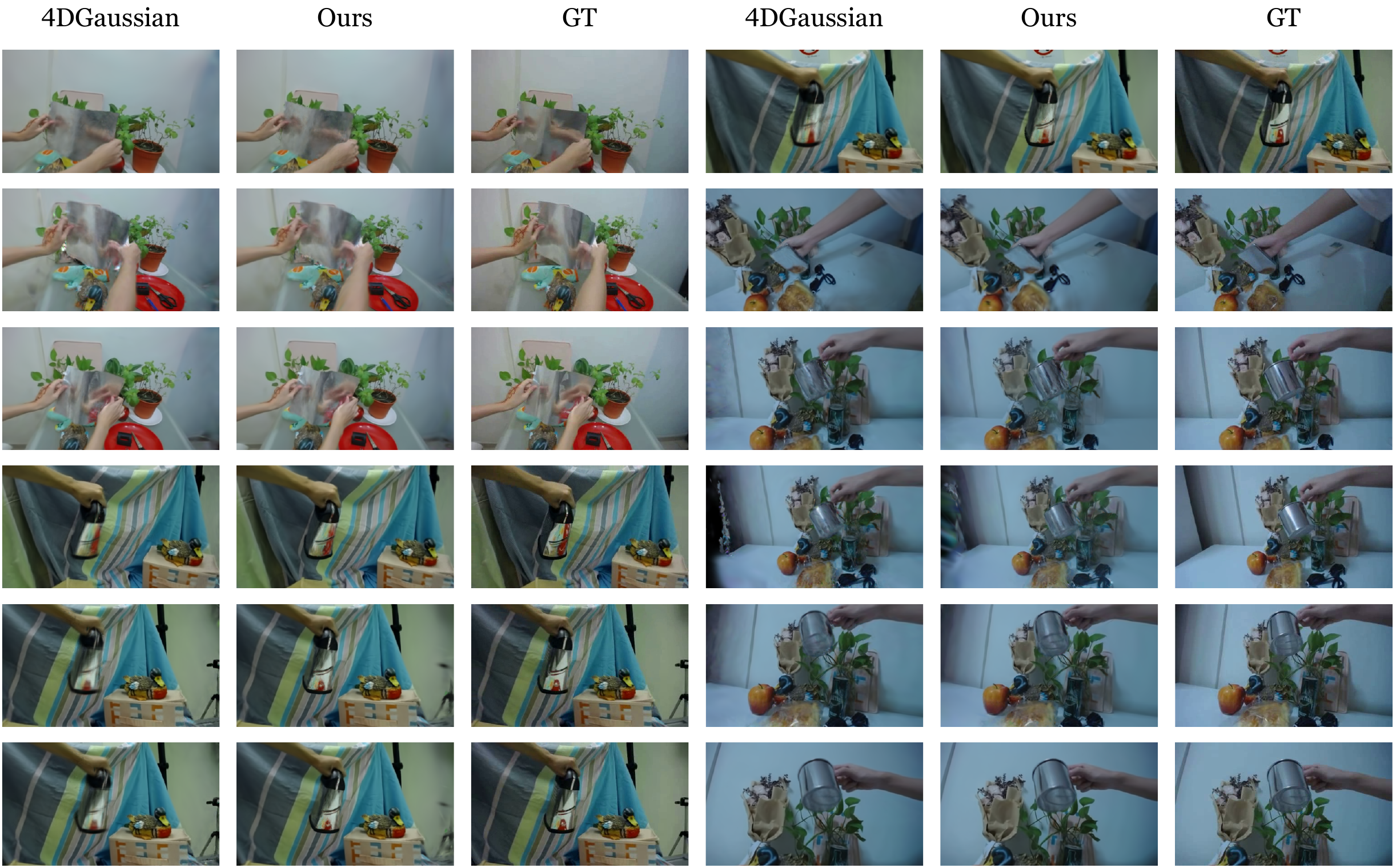}
      \caption{\textbf{Qualitative comparisons on NeRF-DS Dataset.}}
      \label{nerfds}
  \end{center}
  \vspace{-1.3em}
\end{figure}

Figure~\ref{nerfds} shows more visual comparisons on NeRF-DS Dataset. 
We compare the results of 4DGaussian and our model.

The above visual comparisons demonstrate that our method preserves better rendering quality while containing fewer artifacts. 

